\documentclass[10pt,journal,compsoc]{IEEEtran}
\ifCLASSOPTIONcompsoc
  \usepackage[nocompress]{cite}
\else
  \usepackage{cite}
\fi
\ifCLASSINFOpdf
\else
\fi
\usepackage{cite}
\usepackage{amsmath,amssymb,amsfonts}
\usepackage{algorithmic}
\usepackage{graphicx}
\usepackage{textcomp}
\usepackage{xcolor}
\usepackage{multirow}
\usepackage{caption}
\usepackage{mathtools}
\usepackage{booktabs}

\usepackage{algorithm}
\usepackage{listings}

\usepackage{etoolbox}
\makeatletter
\AfterEndEnvironment{algorithm}{\let\@algcomment\relax}
\AtEndEnvironment{algorithm}{\kern2pt\hrule\relax\vskip3pt\@algcomment}
\let\@algcomment\relax
\newcommand\algcomment[1]{\def\@algcomment{\footnotesize#1}}
\renewcommand\fs@ruled{\def\@fs@cfont{\bfseries}\let\@fs@capt\floatc@ruled
  \def\@fs@pre{\hrule height.8pt depth0pt \kern2pt}%
  \def\@fs@post{}%
  \def\@fs@mid{\kern2pt\hrule\kern2pt}%
  \let\@fs@iftopcapt\iftrue}
\makeatother

\newcommand{\etal}{\emph{et al.}}
\newcommand{\eg}{\emph{e.g.}}
\newcommand{\ie}{\emph{i.e.}}
\newcommand{\ProjectName}{CDC}

\begin{document}
%
\title{Contrastive Representation Disentanglement for Clustering}
%
%
%
%

\author{Fei~Ding,
        Dan~Zhang,
        Yin~Yang,~\IEEEmembership{Member,~IEEE,}
        Venkat~Krovi,~\IEEEmembership{Senior Member,~IEEE,}
        and~Feng~Luo,~\IEEEmembership{Senior Member,~IEEE} 
\thanks{Fei~Ding, Dan~Zhang, Yin~Yang and~Feng~Luo are with the School of Computing, Clemson University, Clemson, SC 29634, USA (e-mail: \{feid, dzhang4, yin5, luofeng\}@clemson.edu)}
\thanks{Venkat~Krovi is with the Departments of Automotive Engineering and Mechanical Engineering, Clemson University, Clemson, SC 29634, USA (e-mail: vkrovi@clemson.edu)}
}

\IEEEtitleabstractindextext{%
\begin{abstract}
Clustering continues to be a significant and challenging task. Recent studies have demonstrated impressive results by applying clustering to feature representations acquired through self-supervised learning, particularly on small datasets. However, when dealing with datasets containing a large number of clusters, such as ImageNet, current methods struggle to achieve satisfactory clustering performance. In this paper, we introduce a novel method called Contrastive representation Disentanglement for Clustering (CDC) that leverages contrastive learning to directly disentangle the feature representation for clustering. In CDC, we decompose the representation into two distinct components: one component encodes categorical information under an equipartition constraint, and the other component captures instance-specific factors. To train our model, we propose a contrastive loss that effectively utilizes both components of the representation. We conduct a theoretical analysis of the proposed loss and highlight how it assigns different weights to negative samples during the process of disentangling the feature representation. Further analysis of the gradients reveals that larger weights emphasize a stronger focus on hard negative samples. As a result, the proposed loss exhibits strong expressiveness, enabling efficient disentanglement of categorical information. Through experimental evaluation on various benchmark datasets, our method demonstrates either state-of-the-art or highly competitive clustering performance. Notably, on the complete ImageNet dataset, we achieve an accuracy of 53.4\%, surpassing existing methods by a substantial margin of +10.2\%.
\end{abstract}

}

\maketitle

\IEEEdisplaynontitleabstractindextext

%
\IEEEpeerreviewmaketitle

\section{Introduction}
Image clustering has been widely used in many computer vision, such as image segmentation~\cite{ji2019invariant} and visual features learning~\cite{li2020prototypical,caron2020unsupervised}. Since images are usually high-semantic and high-dimensional, it is difficult to achieve better performance when clustering on large-scale datasets. Earlier clustering studies~\cite{xie2016unsupervised,yang2017towards,hu2017learning,ji2019invariant} focus on end-to-end training solutions. For example, Information Maximizing Self-Augmented Training (IMSAT)~\cite{hu2017learning} and Invariant Information Clustering (IIC)~\cite{ji2019invariant} develop clustering methods from a mutual information maximization perspective, and Deep Embedded Clustering (DEC)~\cite{xie2016unsupervised} and Deep Clustering Network (DCN)~\cite{yang2017towards} perform clustering on initial features obtained from autoencoders. Since these methods rely on network initialization and are likely to focus on low-level non-semantic features~\cite{van2020scan}, such as color and texture, they are prone to cluster degeneracy solutions.

The recently developed clustering methods usually consist of two key steps: representation learning and cluster assignment. Representation learning aims to learn semantically meaningful features, \ie, samples from the same category are projected to similar features so that all samples are linearly separable.
One popular representation learning is self-supervised contrastive learning~\cite{dosovitskiy2015discriminative,wu2018unsupervised,he2020momentum,chen2020simple} that greatly improves the learned representation. To obtain categorical information without labels, an additional step such as K-means clustering~\cite{xie2016unsupervised,caron2018deep,li2020prototypical}  or training of a classifier~\cite{van2020scan} is required for cluster assignment.  K-means clustering~\cite{xie2016unsupervised,caron2018deep,li2020prototypical} is widely used for clustering on learned features. It requires the proper selection of distance measures, thus suffering from the uneven assignment of clusters and leading to a degenerate solution~\cite{caron2018deep}. Semantic Clustering by Adopting Nearest Neighbors (SCAN)~\cite{van2020scan} proposes a novel objective function to train a classifier instead of using K-means. Its performance relies heavily on the feature quality such that nearest neighbors of each sample in the feature space belong to the same category. Due to the presence of noisy nearest neighbors, there is still room for improvement in clustering performance on large-scale datasets.


\begin{figure*}[!t]
\vskip 0.2in
  \centering
  \includegraphics[width=0.7\textwidth]{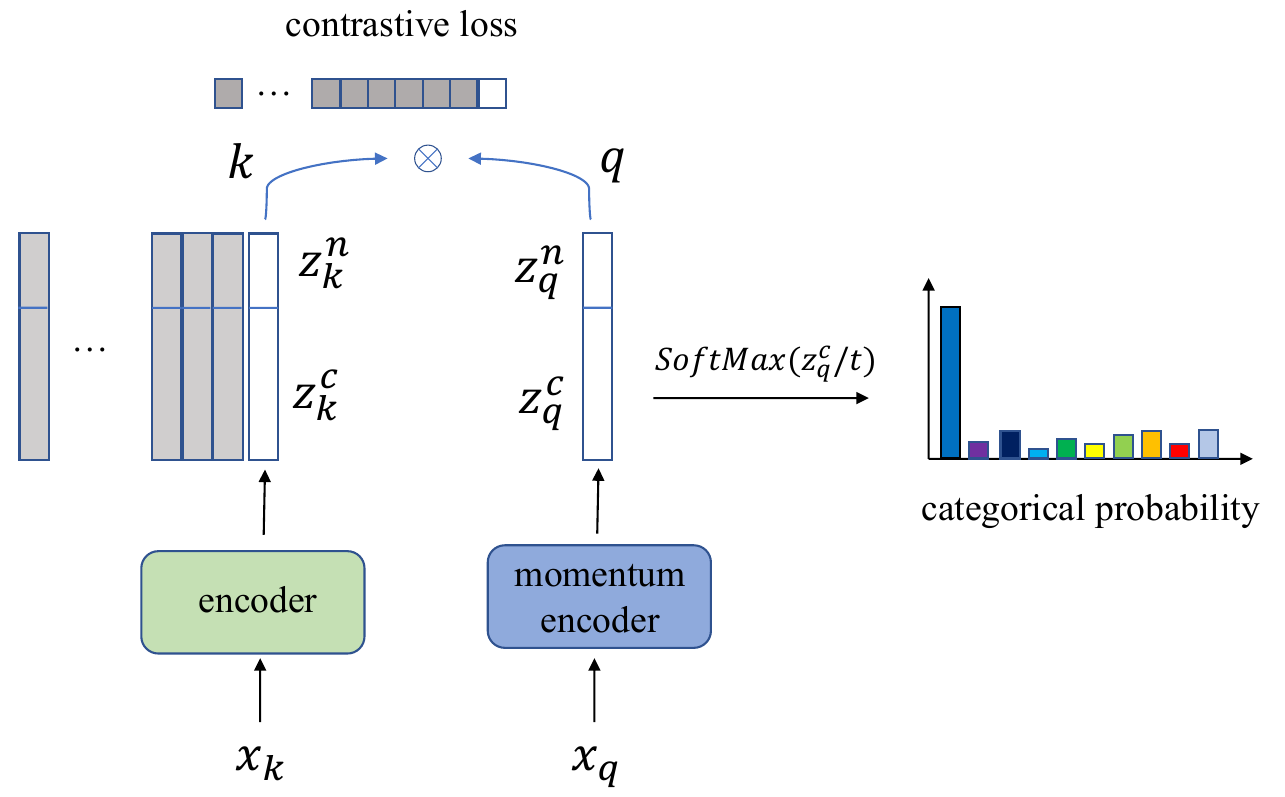}
  \caption{The training framework of~\ProjectName, illustrated by MoCo~\cite{he2020momentum}. The dot product between $q$ and $k$ is written as $q \cdot k = \mathbf{z}_q^n \cdot \mathbf{z}_k^n + \mathbf{z}_q^c \cdot \mathbf{z}_k^c$. After training, we obtain the categorical probability by applying softmax function on $\mathbf{z}^c$. }
  \label{fig:clustering}
\vskip -0.2in
\end{figure*}

In this paper, we propose Contrastive representation Disentanglement based Clustering (\ProjectName), a novel clustering method that directly disentangles the categorical information from the feature representation via contrastive learning. Specifically, we formulate contrastive learning as a proxy task to disentangle the feature representation, which enables us to take advantage of the powerful contrastive learning frameworks. Figure~\ref{fig:clustering} shows an illustration of \ProjectName. First, each representation is decomposed into two parts: $\mathbf{z}^c$ and $\mathbf{z}^n$, where $\mathbf{z}^c$ represents categorical information (logits) and $\mathbf{z}^n$ is used for capturing instance-wise factors. Then, $\mathbf{z}^c$ and $\mathbf{z}^n$ are concatenated together for the training of typical contrastive learning. To avoid the collapse of $\mathbf{z}^c$, we apply the equipartition constraint on it to ensure that the clusters are evenly assigned. We demonstrate that this constraint plays a crucial role in disentangling the representation into two distinct parts and ensuring that $\mathbf{z}^c$ effectively encodes categorical information. By considering $\mathbf{z}^c$ as part of the representation, which can handle well a large number of clusters, we achieve the efficient disentanglement for cluster assignment through contrastive learning.

Our method can be interpreted as instance-wise contrastive learning with self-adjusting weights, which learns to set different weights to distinguish  different negative samples. Typical contrastive learning methods aim to force positive pairs to achieve high similarity (dot product) and negative pairs to achieve low similarity on $\mathbf{z}^n$. Our method adjusts the order of magnitude corresponding to each dimension of $\mathbf{z}^c$ to distinguish between intra-class and inter-class samples. For example, positive pairs yield high similarity scores because they are from the same instance, and negative pairs from different categories yield low similarity scores due to different semantics. Note that negative pairs from the same category yield moderate similarity scores during the instance-wise setting. We demonstrate that the similarity of $\mathbf{z}^c$ provides a mechanism to adjust different weights for negative samples and improves the uniformity property of negative samples in the representation space, therefore beneficial for representation learning. Our contributions can be summarized as follows:
\begin{itemize}
    \item We propose~\ProjectName, a contrastive representation disentanglement-based clustering method that separates categorical information from the feature representation. It considers contrastive learning as a proxy task to efficiently learn disentangled representation. 
    \item We apply the equipartition constraint on part of the representation to enforce representation disentanglement. The proposed objective function plays a key role in disentangling the representation into two parts: categorical and instance-wise information. 
    \item We provide a theoretical analysis to demonstrate that~\ProjectName~adjusts different weights for negative samples through learning cluster assignments. With a gradient analysis, we show that the larger weights tend to concentrate more on hard negative samples.
    \item The clustering experiments show that~\ProjectName~outperforms existing methods in multiple benchmarks, and in particular, achieves significant improvements on ImageNet.~\ProjectName~also contributes to better representation learning results.
\end{itemize}

\section{Method}

Our goal is to disentangle the feature representation for clustering via contrastive learning. There are several representation learning methods~\cite{caron2018deep,asano2019self,li2020prototypical,caron2020unsupervised} that jointly learn clustering and feature representation. For example, SwAV~\cite{caron2020unsupervised} proposes an online codes assignment to learn feature representation by comparing prototypes corresponding to multiple views. These methods either require representation learning and clustering to be performed alternately or require learning additional prototypes, which may be not efficient enough for cluster assignment without labels. It's still necessary to propose a novel objective function to directly obtain cluster assignment. Our method learns cluster assignment via self-adjusting contrastive representation disentanglement and improves the clustering performance on the large-scale dataset. The proposed disentanglement loss can be combined with many contrastive learning methods for cluster assignment. Our method demonstrates that contrastive learning can not only achieve remarkable performance in representation learning, but also high efficiency for representation disentanglement.

\subsection{Weighted Instance-wise Contrastive Learning}
\label{subsec:ICL}

Given a training set $\mathbf{X}=\left\{\mathbf{x}_{1}, \ldots, \mathbf{x}_{N}\right\}$, contrastive learning aims to map $\mathbf{X}$ to $\mathbf{Z}=\left\{\mathbf{z}_{1}, \ldots, \mathbf{z}_{N}\right\}$ with $\mathbf{z}_i = h(f(\mathbf{x}_i))$, such that $\mathbf{z}_i$ can represent $\mathbf{x}_i$ in representation space. $f(\cdot)$ denotes a feature encoder backbone and the projection head $h(\cdot)$ usually is a multi-layer perceptron. Given the similarity $s_{i,j}$, it can be written as: $s_{i, j} = \mathbf{z}_i \cdot \mathbf{z}_j = \mathbf{z}_i^{c} \cdot \mathbf{z}_j^{c} + \mathbf{z}_i^{n} \cdot \mathbf{z}_j^{n}$. Let $s_{i, j}^{c} = \mathbf{z}_i^{c} \cdot \mathbf{z}_j^{c}$, $s_{i, j}^{n} = \mathbf{z}_i^{n} \cdot \mathbf{z}_j^{n}$, then $s_{i, j} = s_{i, j}^{c} + s_{i, j}^{n}$ and $\exp (s_{i, j} / \tau) = \exp (s_{i, j}^{c} / \tau) \cdot \exp (s_{i, j}^{n} / \tau)$. The objective function of contrastive learning, such as InfoNCE~\cite{van2018representation,chen2020simple,he2020momentum}, can be formulated as:
\begin{align}
& \mathcal{L}_{\operatorname{InfoNCE}}\left(\mathbf{x}_{i}\right) \\ &=-\log \frac{\exp \left(s_{i, i} / \tau\right)}{\sum_{k} \exp \left(s_{i, k} / \tau\right)+\exp \left(s_{i, i} / \tau\right)} = \label{eq:infonce}\\
& -\log \frac{\exp ( s_{i, i}^{n} / \tau )}{\sum_{k} (\exp ( ( s_{i, k}^{c} - s_{i, i}^{c} ) / \tau ) \cdot \exp (s_{i, k}^{n} / \tau ) )+\exp (s_{i, i}^{n} / \tau )}, \label{eq:infonce_zczn}
\end{align}
where $\tau$ is a temperature hyper-parameter, the positive similarity $s_{i, i}$ is calculated by two augmented versions of the same image, and the negative similarity $s_{i,j} (j \neq i)$ compares different images. In our settings, $\mathbf{z}_i$ consists of $\mathbf{z}_i^{c}$ and $\mathbf{z}_i^{n}$. The similarity $s_{i,j}$ can be written as: $s_{i, j} = \mathbf{z}_i \cdot \mathbf{z}_j = \mathbf{z}_i^{c} \cdot \mathbf{z}_j^{c} + \mathbf{z}_i^{n} \cdot \mathbf{z}_j^{n}$. Let $s_{i, j}^{c} = \mathbf{z}_i^{c} \cdot \mathbf{z}_j^{c}$, $s_{i, j}^{n} = \mathbf{z}_i^{n} \cdot \mathbf{z}_j^{n}$, then we can re-write the standard contrastive loss (equation~\ref{eq:infonce}) as equation~\ref{eq:infonce_zczn}. 

Note that $s_{i, i}^{n}$ can be considered as instance-wise positive similarity, $s_{i, k}^{n}$ can be considered as instance-wise negative similarity, and $\exp ( ( s_{i, k}^{c} - s_{i, i}^{c} ) / \tau )$ is a learnable coefficient for each negative sample. Thus, we obtain a more expressive contrastive loss. Considering that $\tau$ is a hyperparameter, the value of this coefficient is mainly determined by $s_{i, k}^{c}$ and $s_{i, i}^{c}$, which are further calculated by $\mathbf{z}_i^c$ and $\mathbf{z}_k^c$. This coefficient learns to set different weights for each negative sample, \ie, larger weight on hard negative (intra-class) samples and smaller weight on inter-class negative samples. In this way, the part of representation $\mathbf{z}^c$ plays a role in adjusting different penalties on different negative samples. In contrast, the typical contrastive loss sets all negative samples to the same coefficient with a value of 1.

Since $s_{i, j}^{c}$ and $s_{i, j}^{n}$ are symmetric in equation~\ref{eq:infonce}, the standard contrastive loss can also be written as:
\begin{align}
& \mathcal{L}_{\operatorname{InfoNCE}}\left(\mathbf{x}_{i}\right) =-\log \frac{\exp \left( s_{i, i} / \tau \right)}{\sum_{k} \exp \left( s_{i, k} / \tau \right)+\exp \left( s_{i, i} / \tau \right)}\\
& = -\log \frac{\exp ( s_{i, i}^{c} / \tau )}{\sum_{k} (\exp ( (s_{i, k}^{n} - s_{i, i}^{n}) / {\tau} ) \cdot \exp ( s_{i, k}^{c} / {\tau} ) )+\exp (s_{i, i}^{c} / \tau)}. \label{eq:infonce_znzc}
\end{align}

Then we observe that the coefficient $\exp ( ( s_{i, k}^{n} - s_{i, i}^{n} ) / \tau )$ can be considered as a constant because $\mathbf{z}^n$ satisfies the properties of alignment and uniformity, as analyzed in Section~\ref{subsec:latent}. Thus, equation~\ref{eq:infonce_znzc} can be considered as a standard contrastive loss on $\mathbf{z}^c$. $\mathbf{z}^c$ can retain well the categorical information ( shown in Figure~\ref{fig:zn_visualization} (a)), which is beneficial for learning cluster assignment via contrastive learning.

To make the above coefficient work as expected, we need to add some constraints to $\mathbf{z}^c$. First, the constraint should ensure that the value range of coefficients is bounded. This requirement can be satisfied by performing normalization on $\mathbf{z}^c$ and $\mathbf{z}^n$ separately. More importantly, the constraint should satisfy that $\mathbf{z}^c$ does not collapse to the same assignment, otherwise our method will degrade to typical contrastive learning. Here, we introduce the equipartition constraint on $\mathbf{z}^c$ which encourages it to encode the semantic structural information, while also avoiding its collapse problem. $\mathbf{z}^c$ is expected to represent the probability over clusters $\mathcal{C}=\{1, \ldots, K\}$ after softmax function.

\subsection{Equipartition constraint for disentanglement}

Given the logits $\mathbf{z}^c$, we can obtain the categorical probabilities via the softmax function:  $\mathbf{p}\left(y \mid \mathbf{x}_{i}\right)=\operatorname{softmax}\left(\mathbf{z}_i^c / t \right)$, where $t$ is the temperature to rescale the logits score. To avoid degeneracy, we add the equipartition constraint on cluster assignment to enforce that the clusters are evenly assigned in a batch. The pseudo-assignment $\mathbf{q}\left(y \mid \mathbf{x}_{i}\right) \in\{0,1\}$ is used to describe the even assignment of $\mathbf{z}^c$. We denote $B$ logits in a batch by $\mathbf{Z}^{c}=\left[\mathbf{z}_{1}^{c} / t, \ldots, \mathbf{z}_{B}^{c} / t \right]$, and the pseudo-assignment by $\mathbf{Q}=\left[\mathbf{q}_{1}, \ldots, \mathbf{q}_{B}\right]$.

The equipartition constraint has been used in previous self-supervised learning studies~\cite{asano2019self,caron2020unsupervised} for representation learning. Asano~\etal~\cite{asano2019self} propose to solve the matrix $\mathbf{Q}$ by restricting the transportation polytope on the entire training dataset. SwAV~\cite{caron2020unsupervised} improves the above solution to calculate the online prototypes for contrastive learning, and achieves excellent representation learning results. Unlike SwAV, which uses the similarity of features and prototypes as input to obtain pseudo-assignment, whose assignment can be interpreted as the probability of assigning each feature to a prototype, we consider the representation $\mathbf{z}^c$ as logits to directly obtain assignments without any prototypes. Here, we propose to adopt a similar solution to optimize $\mathbf{Q}$ directly from the logits matrix $\mathbf{Z}^{c}$, 
\begin{equation}
\label{eq:sk}
\max _{\mathbf{Q} \in \mathcal{Q}} \operatorname{Tr}\left(\mathbf{Q}^{\top} \mathbf{Z}^{c}\right)+\varepsilon H(\mathbf{Q}),
\end{equation}
where $\mathcal{Q}=\left\{\mathbf{Q} \in \mathbb{R}_{+}^{K \times B} \mid \mathbf{Q} \mathbf{1}_{B}=\frac{1}{K} \mathbf{1}_{K}, \mathbf{Q}^{\top} \mathbf{1}_{K}=\frac{1}{B} \mathbf{1}_{B}\right\}$ and H denotes the entropy regularization. $\mathbf{1}_{B}$ and $\mathbf{1}_{K}$ denote the vector of ones in dimension B and K. These constraints ensure that all samples in each batch are evenly divided into K clusters. We also set $\varepsilon$ to be small to avoid a trivial solution~\cite{caron2020unsupervised}. 

The solution of equation~\ref{eq:sk} can be written as: $\mathbf{Q}^{*}=\operatorname{Diag}(\mathbf{u}) \exp \left(\frac{\mathbf{Z}^{c}}{\varepsilon}\right) \operatorname{Diag}(\mathbf{v})$, where $\mathbf{u}$ and $\mathbf{v}$ are two scaling vectors such that $\mathbf{Q}$ is a probability matrix~\cite{asano2019self,cuturi2013sinkhorn}. The vectors $\mathbf{u}$ and $\mathbf{v}$ can be computed using Sinkhorn-Knopp algorithm~\cite{cuturi2013sinkhorn} through several iterations. In practice, by using GPU, 3 iterations are fast enough and can ensure satisfactory results~\cite{caron2020unsupervised}. Once we obtain the solution $\mathbf{Q}^{*}$, we directly apply its soft assignment to constrain $\mathbf{z}^{c}$ by minimizing the following cross-entropy loss for disentanglement:
\begin{equation}
\label{eq:ce}
\mathcal{L}_{\operatorname{CE}}\left(\mathbf{z}^{c}, \mathbf{q}\right)=-\sum_{k} \mathbf{q}^{(k)} \log \mathbf{p}^{(k)}.
\end{equation}


\subsection{Gradients Analysis}

Here, we perform a gradient analysis to understand the properties of the proposed contrastive loss. Because the equipartition constraint is not related to negative similarity $s_{i, j}^{n} (j \neq i)$, for convenience, our analysis focuses on the negative gradients. Considering that the magnitude of positive gradient $\frac{\partial \mathcal{L}\left(\mathbf{x}_{i}\right)}{\partial s_{i, i}^{n}}$ is equal to the sum of all negative gradients, we can also indirectly understand the property of the positive gradient through negative gradients. The gradient with respect to the negative similarity $s_{i, j}^{n} (j \neq i)$ is formulated as: 
\begin{equation}
\label{eq:gradients}
\frac{\partial \mathcal{L}\left(\mathbf{x}_{i}\right)}{\partial s_{i, j}^{n}}=\frac{\lambda_{i, j}^{c}}{\tau} \cdot \frac{\exp ( s_{i, j}^{n} / \tau )}{\sum_{k \neq i} (\lambda_{i, k}^{c} \cdot \exp (s_{i, k}^{n} / \tau ) )+\exp (s_{i, i}^{n} / \tau )},
\end{equation}
where $\lambda_{i, j}^{c} = \exp ( ( s_{i, j}^{c} - s_{i, i}^{c} ) / \tau)$. Without the loss of generality, the hyperparameter $\tau$ can be considered as a constant.

From equation~\ref{eq:gradients}, we observe that $\lambda_{i, j}^{c}$ is proportional to negative gradients. A larger $\lambda_{i, j}^{c}$ leads to the corresponding sample to receive more attention during the optimization. Since $\lambda_{i, j}^{c}$ depends mainly on $\mathbf{z}_i^c$ and $\mathbf{z}_j^c$, we need to analyze them separately according to whether samples belong to the same category or not. Due to the equipartition constraint, $\mathbf{z}^c$ is encouraged to encode the categorical information. Thus, the similarity $s_{i,j}^c (j \neq i)$ of the same category is greater than the similarity of different categories. The intra-class $\lambda_{i, j}^{c}$ is also greater than the inter-class $\lambda_{i, j}^{c}$. In other words, the gradient tends to concentrate more on samples of the same category, which are often considered as hard negative samples. In this way, the categorical information of $\mathbf{z}^c$ can contribute to the optimization of $\mathbf{z}^n$ so that all samples tend to be uniformly distributed.

\subsection{\ProjectName~objective}

Our overall objective, namely~\ProjectName, is defined as
\begin{equation}
\label{eq:objective}
    \mathcal{L} = \mathcal{L}_{\operatorname{InfoNCE}} + \alpha \mathcal{L}_{\operatorname{CE}}.
\end{equation}
In addition to the loss weight $\alpha$, there are two temperature hyperparameters: $\tau$ and $t$. We observe that the choice of temperature values has a crucial impact on the clustering performance. In general, the relationship of temperatures satisfies: $0 < t \le \tau \le 1$. We refer to Section~\ref{subsec:temperature} for the concrete analysis.

Because samples with the highly confident prediction (close to 1) can be considered to obtain pseudo labels, our method can optionally include the confidence-based cross-entropy loss~\cite{van2020scan}, which can be gradually added to the overall objective, or be used for fine-tuning on the pretrained model. SCAN applies this loss to correct for errors introduced by noisy nearest neighbors, while we aim to encourage the model to produce a smooth feature space, thus helping assign proper clusters for boundary samples. We only consider well-classified samples, \ie, $p_{\operatorname{max}} > $ threshold (0.99), and perform strong data augmentation on them. This encourages different augmented samples to output consistent cluster predictions through the cross-entropy loss, also known as self-labeling~\cite{van2020scan}.

Algorithm~\ref{alg:code} provides PyTorch-like pseudo-code to describe how we compute the objective (equation~\ref{eq:objective}).

\begin{algorithm}[t]
\caption{PyTorch-like Pseudo-code for \ProjectName.}
\label{alg:code}
\algcomment{\fontsize{7.2pt}{0em}\selectfont \texttt{Softmax}: Softmax function; \texttt{cat}: concatenation.
}
\definecolor{codeblue}{rgb}{0.25,0.5,0.5}
\lstset{
  backgroundcolor=\color{white},
  basicstyle=\fontsize{7.2pt}{7.2pt}\ttfamily\selectfont,
  columns=fullflexible,
  breaklines=true,
  captionpos=b,
  commentstyle=\fontsize{7.2pt}{7.2pt}\color{codeblue},
  keywordstyle=\fontsize{7.2pt}{7.2pt},
}
\begin{lstlisting}[language=python]
# model: includes base_encoder, momentum_encoder and MLP heads
# T and t: temperatures for contrastive loss and cross entropy loss
# alpha: weight of the loss term
# K: dimension of zc (number of clusters), C: dimension of zn

for x in loader:  # load a minibatch x with N samples
    x_q = aug(x) # a randomly augmented version
    x_k = aug(x) # another randomly augmented version
    
    # no gradient to k
    q, k = model.forward(x_q, x_k)  # compute features: N x (K + C)

    zc_q = normalize(q[:, :K], dim=1) # normalize zc: N x K
    zn_q = normalize(q[:, K:], dim=1) # normalize zn: N x C
    q = cat([zc_q, zn_q], dim=1)

    zc_k = normalize(k[:, :K], dim=1) # normalize zc: N x K
    zn_k = normalize(k[:, K:], dim=1) # normalize zn: N x C
    k = cat([zc_k, zn_k], dim=1)  
    
    # compute assignments with sinkhorn-knopp
    with torch.no_grad():
        q_q = sinkhorn-knopp(zc_q)
        q_k = sinkhorn-knopp(zc_k)
    
    # convert logits to probabilities
    p_q = Softmax(zc_q / t)
    p_k = Softmax(zc_k / t)
    
    # compute the equipartition constraint
    cross_entropy_loss = - 0.5 * mean(q_q * log(p_k) + q_k * log(p_q))
    
    loss = contrastive_loss(q, k, T) + alpha * cross_entropy_loss

    # SGD update: network and MLP heads
    loss.backward()
    update(model.params)

\end{lstlisting}
\end{algorithm}

\section{Experiments}
\label{section:experiments}

In this section, we evaluate \ProjectName~on multiple benchmarks, including training models from scratch and using self-supervised pretrained models. We follow the settings in MoCo~\cite{he2020momentum} and choose the same backbone network as the baseline methods, to ensure that our performance gains are from the proposed objective function. We first compare our results to the state-of-the-art clustering methods, where we find that our method is overall the best or highly competitive in many benchmarks. Then, we quantify the representation learned by the proposed contrastive loss, and the results show that it can also improve the representation quality.

\subsection{Experimental setup}

\textbf{Datasets.} We perform the experimental evaluation on CIFAR10~\cite{krizhevsky2009learning}, CIFAR100-20~\cite{krizhevsky2009learning}, STL10~\cite{coates2011analysis} and ImageNet~\cite{deng2009imagenet}. Some prior works~\cite{xie2016unsupervised,ji2019invariant,chang2017deep} use the full dataset for both training and evaluation. Here, we follow the experimental settings in SCAN~\cite{van2020scan}, which trains and evaluates the model using train and val split respectively. This helps us to understand the generalizability of our method on unseen samples. All datasets are processed using the same augmentations in MoCo v2~\cite{chen2020mocov2}. We report the results with the mean and standard deviation from 5 different runs.

\noindent
\textbf{Implementation details.} We apply the standard ResNet~\cite{he2016deep} backbones (ResNet-18 and ResNet-50) each with a MLP projection head. The dimensionality of $\mathbf{z}^c$ is determined by the number of clusters, and the dimensionality of $\mathbf{z}^n$ is set to 256 on the ImageNet and 128 on the other datasets. 

On the smaller datasets (CIFAR10, CIFAR100-20 and STL10), our implementation is based on the Lightly library~\cite{susmelj2020lightly}. The parameters are trained through the SGD optimizer with a learning rate of 6e-2, a momentum of 0.9, and a weight decay of 5e-4. The loss term is set to $\alpha = 5.0$, and two temperature values are set to $\tau = 0.15$ and $t = 0.10$. We adopt the same data augmentation in SimCLR~\cite{chen2020simple} but disable the blur like MoCo v2~\cite{chen2020mocov2}. We train the models from scratch for 1200 epochs using batches of size 512. For the linear classification in the ablation studies,
we train the linear classifier via the SGD optimizer with a learning rate of 30 and the cosine scheduler for 100 epochs. 

For ImageNet subsets (100 classes and 200 classes), we adopt the implementation from MoCo v2 and follow the same data augmentation settings in MoCo v2. To speed up training, we directly initialize the backbone with the released pretrained weights (800 epochs pretrained) like SCAN, and only train the MLP head. The weights are updated through SGD optimizer with a learning rate of 0.03, a momentum of 0.9, and a weight decay of 1e-4. The three hyperparameters are set to $\alpha = 1.0$, $\tau = 0.40$ and $t = 0.20$. We train the network weights for 400 epochs using a batch size of 256. Although there are advanced data augmentation and training strategies, we adopt the same settings in MoCo v2 for a fair comparison.

\noindent
\textbf{Equipartition constraint.} Most of the Sinkhorn-Knopp implementation are directly from SwAV work~\cite{caron2020unsupervised}. The regularization parameter is set to $\epsilon = 0.05$ and the number of Sinkhorn iterations is set to 3 for all experiments.

\subsection{Comparison with state-of-the-art methods}
\label{subsec:smaller_dataset}

\setlength{\tabcolsep}{4pt}
\begin{table*}[t]
\begin{center}
\caption{State-of-the-art comparison: We report the averaged results (Avg $\pm$ Std) for 5 different runs after the clustering and self-labeling steps. All the baseline results are from~\cite{van2020scan}. We train and evaluate the model using the train and val split respectively, which is consistent with the SCAN~\cite{van2020scan}.}
\label{table:sota}
\begin{tabular}{@{}l c ccc c ccc c ccc@{}}
\hline
\textbf{Dataset} && \multicolumn{3}{c}{\textbf{CIFAR10}} && \multicolumn{3}{c}{\textbf{CIFAR100-20}} && \multicolumn{3}{c}{\textbf{STL10}} \\
\cmidrule{3-5} \cmidrule{7-9} \cmidrule{11-13}
\textbf{Metric} && ACC & NMI & ARI && ACC & NMI & ARI&& ACC & NMI & ARI \\
\hline

K-means~\cite{wang2014optimized}     && 22.9 & 8.7   & 4.9  && 13.0  & 8.4  & 2.8 && 19.2 & 12.5 & 6.1  \\
SC~\cite{zelnik2005self}             && 24.7 & 10.3  & 8.5  && 13.6  & 9.0  & 2.2 && 15.9 & 9.8  & 4.8  \\
Triplets~\cite{schultz2003learning}  && 20.5 & --    & --   && 9.94  & --   & --  && 24.4 & --   & --   \\
JULE~\cite{yang2016joint}            && 27.2 & 19.2  & 13.8 && 13.7  & 10.3 & 3.3 && 27.7 & 18.2 & 16.4 \\
AEVB~\cite{kingma2013auto}           && 29.1 & 24.5  & 16.8 && 15.2  & 10.8 & 4.0 && 28.2 & 20.0 & 14.6 \\
DAE~\cite{vincent2010stacked}        && 29.7 & 25.1  & 16.3 && 15.1  & 11.1 & 4.6 && 30.2 & 22.4 & 15.2 \\
SWWAE~\cite{zhao2015stacked}         && 28.4 & 23.3  & 16.4 && 14.7  & 10.3 & 3.9 && 27.0 & 19.6 & 13.6 \\
AE~\cite{bengio2007greedy}           && 31.4 & 23.4  & 16.9 && 16.5  & 10.0 & 4.7 && 30.3 & 25.0 & 16.1 \\
GAN~\cite{radford2015unsupervised}   && 31.5 & 26.5  & 17.6 && 15.1  & 12.0 & 4.5 && 29.8 & 21.0 & 13.9 \\
DEC~\cite{xie2016unsupervised}                       && 30.1 & 25.7  & 16.1 && 18.5  & 13.6 & 5.0 && 35.9 & 27.6 & 18.6 \\
ADC~\cite{haeusser2018associative}   && 32.5 & --    & --   && 16.0  & --   & --  && 53.0 & --   & --   \\
DeepCluster~\cite{caron2018deep}       && 37.4 & --    & --   && 18.9  & --   & --  && 33.4 & --   & --   \\
DAC~\cite{chang2017deep}                       && 52.2 & 40.0 & 30.1 && 23.8& 18.5 & 8.8 && 47.0 & 36.6 & 25.6 \\
IIC~\cite{ji2019invariant} && 61.7 & 51.1  & 41.1 && 25.7  & 22.5 & 11.7 && 59.6 & 49.6 & 39.7 \\
Pretext~\cite{chen2020simple} + K-means && $65.9\pm5.7$ & $59.8\pm2.0$ & $50.9\pm3.7$   && $39.5\pm1.9$ & $40.2\pm1.1$ & $23.9\pm1.1$ && $65.8\pm5.1$ & $60.4\pm2.5$ & $50.6\pm4.1$\\
SCAN~\cite{van2020scan}  &&$81.8\pm0.3$ & $71.2\pm 0.4$ & $66.5\pm0.4 $&& $42.2\pm3.0$ & $44.1\pm1.0$ & $26.7\pm1.3$ && $75.5\pm2.0$ & $65.4\pm1.2$&$59.0\pm1.6$  \\
SCAN~\cite{van2020scan} + selflabel && \underline{$87.6\pm0.4$} & \underline{$78.7\pm 0.5$} & \underline{$75.8\pm0.7 $} && \underline{$45.9\pm2.7$} & \underline{$46.8\pm1.3$} & \underline{$30.1\pm2.1$} && \textbf{76.7 $\pm$ 1.9} & \textbf{68.0 $\pm$ 1.2} & \textbf{61.6 $\pm$ 1.8}  \\
\hline
Supervised  && 93.8 &  86.2 & 87.0   && 80.0 & 68.0 & 63.2 && 80.6 & 65.9 & 63.1\\
\textbf{\ProjectName}  && $83.1\pm 0.6$ & $73.4 \pm 0.7$ & $68.7 \pm 0.6$ && $44.0 \pm 1.2$ & $46.3 \pm 1.1$ & $28.2 \pm 1.0$ && 71.2 $\pm$ 1.5 & 64.3 $\pm$ 1.3 & 55.3 $\pm$ 1.5 \\
\textbf{\ProjectName} + selflabel && \textbf{89.0 $\pm$ 0.3} & \textbf{80.6 $\pm$ 0.4} & \textbf{78.4 $\pm$ 0.5} && \textbf{49.2 $\pm$ 0.8} & \textbf{50.4 $\pm$ 0.5} & \textbf{34.6 $\pm$ 0.6} && 75.2 $\pm$ 0.8 & 66.8 $\pm$ 0.5 & 59.4 $\pm$ 0.6 \\
\hline
\end{tabular}
\end{center}
\end{table*}
\setlength{\tabcolsep}{1.4pt}

We first evaluate \ProjectName's clustering performance on three different benchmarks. We report the results of clustering accuracy (ACC), normalized mutual information (NMI) and adjusted rand index (ARI) in Table~\ref{table:sota}. \ProjectName~outperforms other clustering methods in two of the benchmarks and is on par with state-of-the-art performance in another benchmark. Our method further reduces the gap between clustering and supervised learning on CIFAR-10. On CIFAR100-20, the reason why there is still a large gap with supervised learning is due to the ambiguity caused by the superclasses (mapping 100 classes to 20 classes). On the STL10 dataset, we train the model on the train+unlabeled split from scratch due to the small size of the train split. However, the exact number of clusters in the train+unlabeled split is unknown. We choose the number of clusters equal to 10 for the evaluation on the test dataset. Nevertheless, our method still achieves competitive clustering results, which demonstrates that our method is also applicable to datasets with an unknown number of clusters. Note that the results of other state-of-the-art methods are based on the pretrained representations from contrastive learning, while our clustering results are obtained by training the model from scratch in an end-to-end manner. This also confirms that $\mathbf{z}^c$ efficiently encodes the categorical information. In Section~\ref{subsec:linear}, we further verify that the presence of $\mathbf{z}^c$ enables us to learn better feature representation.

\subsection{ImageNet clustering}

\textbf{ImageNet - subset.} We first test our method on ImageNet subsets of 100 and 200 classes, which is consistent with SCAN. All compared methods apply the same pre-trained weights from MoCo v2~\cite{chen2020mocov2}. We fix the ResNet-50 (R50) backbone and train the MLP projection head for 400 epochs with the same settings as MoCo v2. Table~\ref{table: imagenet_subsets} shows that our method can further improve the clustering accuracy, where the clustering results of K-means and SCAN are from SCAN. For example, it has achieved a much higher accuracy of 61.4\% than SCAN (56.3\%) on the subset of 200 classes. This also implies that~\ProjectName~is applicable to large-scale datasets with a large number of clusters.


\setlength{\tabcolsep}{4pt}
\begin{table}
\begin{center}
\caption{Comparison with other clustering methods on the full ImageNet dataset (1000 classes). We obtain the clustering results on the MoCo V3 pretrained weights by using the code provided by SCAN (*). All compared methods are based on the ResNet-50 backbone. }
\label{table: imagenet}
\scriptsize
\begin{tabular}{@{}l l ccc@{}}
\toprule
\textbf{Method} & \textbf{Backbone} & \textbf{ACC} & \textbf{NMI} & \textbf{ARI} \\
\midrule
MoCo V2 + SCAN~\cite{van2020scan} & R50 & 39.9 & 72.0 & 27.5 \\
MoCo V3 + SCAN{*} & R50 & 43.2 & 70.9 & - \\
MoCo V3 + \ProjectName~ & R50 & \textbf{53.4} & \textbf{76.3} & \textbf{34.7} \\
\bottomrule
\end{tabular}
\end{center}
\end{table}
\setlength{\tabcolsep}{1.4pt}

\begin{figure*}[t]
\begin{minipage}[t]{0.48\linewidth}
\centering
\includegraphics[width=1.00\textwidth]{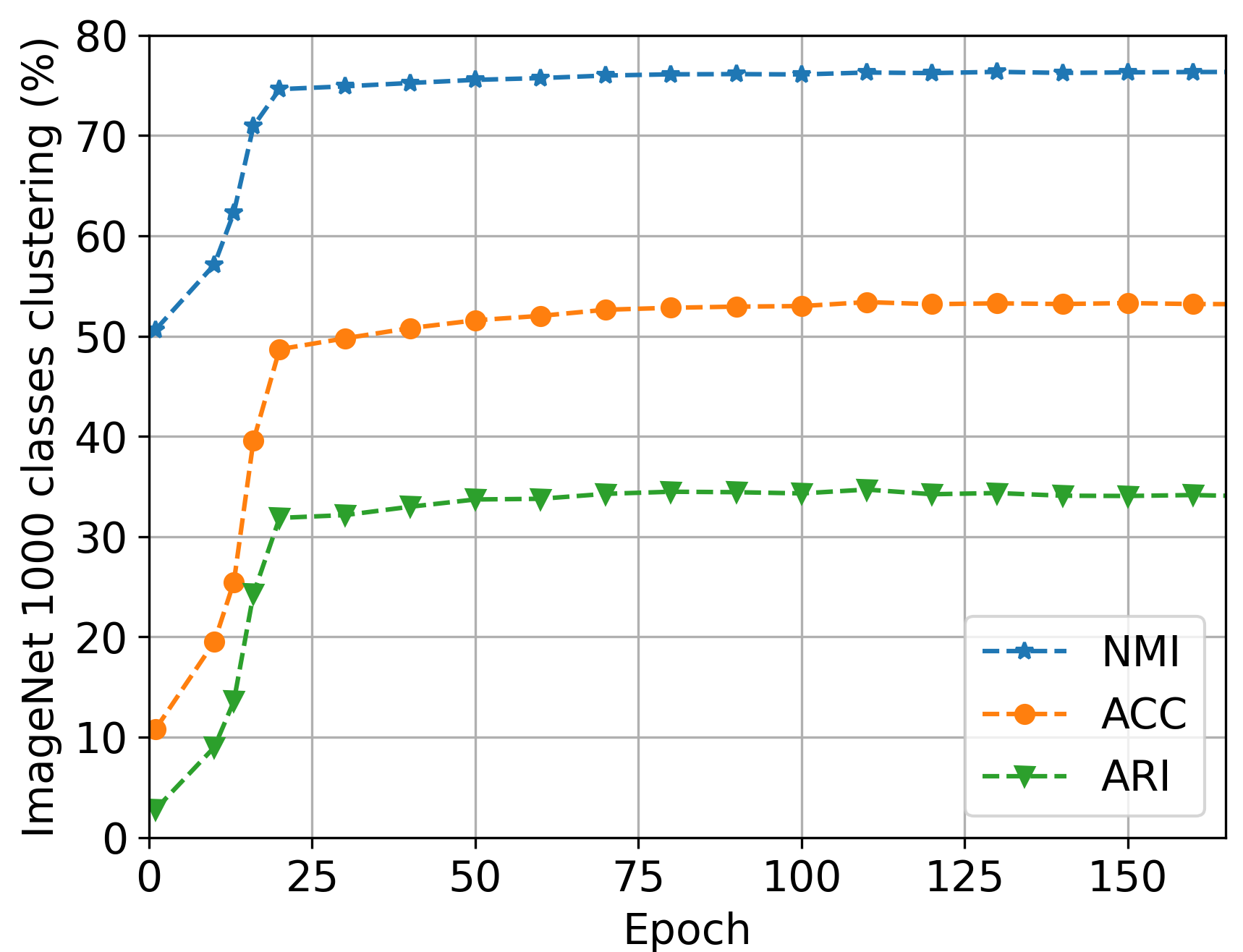}
\caption{Clustering results of \ProjectName~($t = 0.1$) on full ImageNet dataset (1000 classes) up to 200 epochs.}
\label{fig:epoch_clustering}
\end{minipage}
\hspace*{\fill}
\begin{minipage}[t]{0.48\linewidth}
\centering
\includegraphics[width=1.00\textwidth]{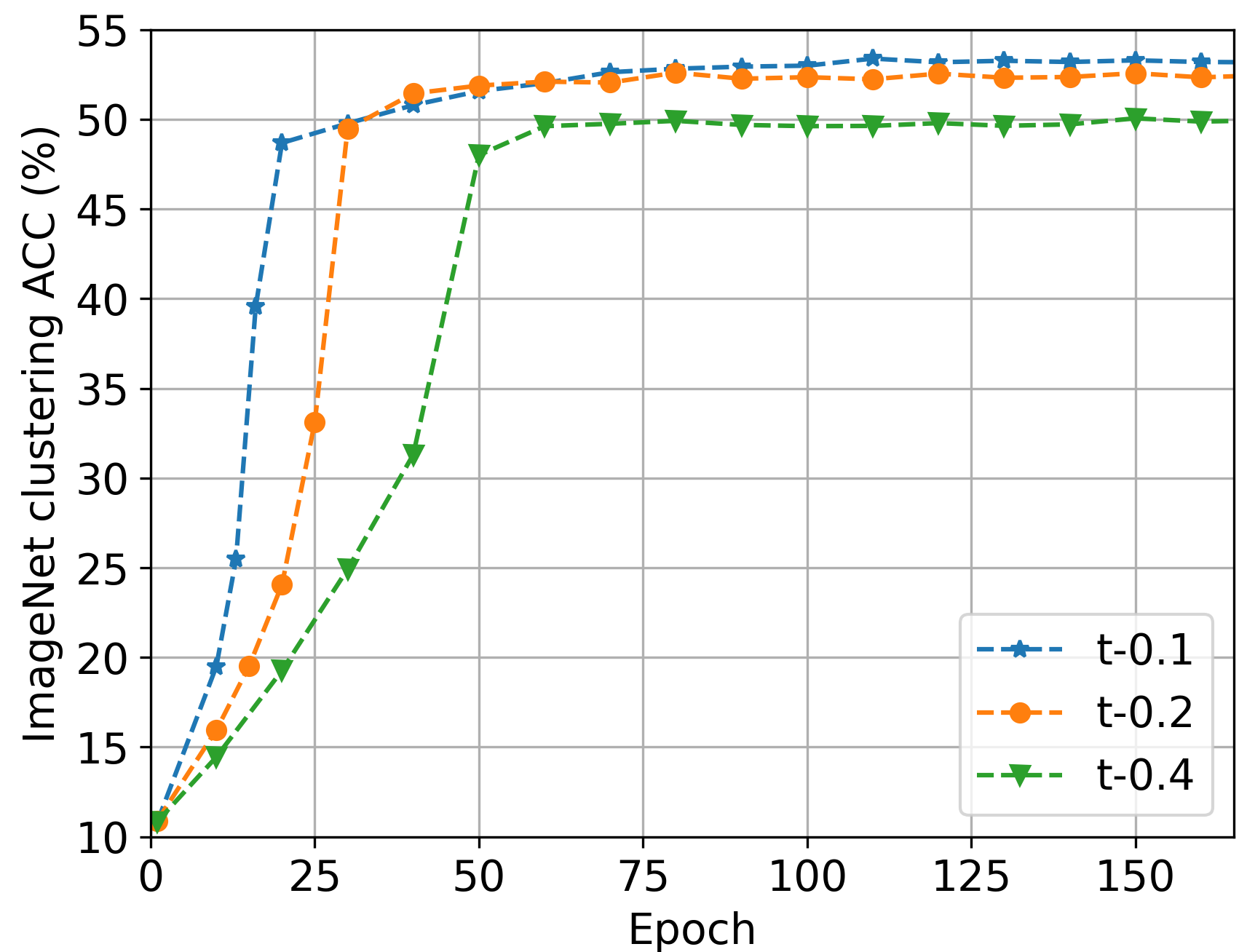}
\caption{Comparison of clustering accuracy at different temperatures on full ImageNet dataset.}
\label{fig:temperature_clustering}
\end{minipage}
\end{figure*}

\noindent
\textbf{ImageNet - full.} We further consider the clustering evaluation on the full ImageNet dataset. We apply our method to the latest contrastive learning studies, such as MoCo v3~\cite{chen2021mocov3}, to uncover its potential in clustering tasks. We load the pretrained R50 backbone from MoCo v3 (1000 epochs) and only train two MLP heads for 200 epochs with the same settings as MoCo V3. We use the LARS optimizer with a learning rate of 0.3, a weight decay of 1e-6 and a momentum of 0.9.

Table~\ref{table: imagenet} compares our method against SCAN on three metrics. \ProjectName~consistently outperforms the baseline method in all metrics. In particular, it achieves significant performance improvements in terms of accuracy (53.4\%) compared to SCAN (43.2\%). Although previous studies~\cite{van2020scan} find that there may be multiple reasonable ways to cluster images in ImageNet based on their semantics, without a priori knowledge, it's still challenging to cluster images in ImageNet according to their true labels. But~\ProjectName~still achieves promising clustering results, which demonstrates the advantages of the proposed method. Figure~\ref{fig:epoch_clustering} shows the training efficiency of our method, which can converge in a small number of epochs.

\subsection{Linear evaluation}\label{subsec:linear}

We follow the same settings as MoCo v2 to enable a reasonable evaluation of the benefits due to the introduction of $\mathbf{z}^c$. We perform the same data augmentation and training strategy to train the model on ImageNet training data for 200 epochs from scratch. Then, we fix the R50 backbone and train a linear classifier to evaluate the learned feature encoder on three datasets: ImageNet, VOC07~\cite{everingham2010pascal}, and Places205~\cite{zhou2014learning}. Table~\ref{table: imagenet_linear} shows that the proposed contrastive objective achieves competitive results on these linear classification tasks. Especially, for the transfer learning on VOC07 and Places205, it demonstrates that our method achieves better generalizability to the downstream tasks than other methods. PCL v2~\cite{li2020prototypical} is another method that utilizes clustering to improve representation learning. Although our main purpose of introducing $\mathbf{z}^c$ is for clustering, it can also improve the quality of feature representation. This demonstrates that the expressiveness of the model is improved by the negative coefficients due to the introduction of $\mathbf{z}^c$. We continue to evaluate the pretrained ResNet-50 backbone on object detection tasks in the following Section.

\setlength{\tabcolsep}{4pt}
\begin{table}[t]
\begin{center}
\caption{Image classification with linear classifiers. We report the top-1 accuracy for ImageNet and Places205, mAP for VOC dataset. All the baseline results are from~\cite{li2020prototypical} and \cite{caron2020unsupervised}. Our results are obtained by directly applying the evaluation code~\cite{hu2021adco} on the pretrained R50 backbone.}
\label{table: imagenet_linear}
\begin{tabular}{@{}l l l c c c}
\toprule
    \multirow{2}{*}{Method}	 & \multirow{2}{*}{Architecture} &\#pretrain  &\multicolumn{3}{c}{Dataset}\\
	& &epochs& ImageNet& VOC07 & Places205 \\
\midrule
\noalign{\smallskip}
    Supervised & R50 & - & 76.5 & 87.5 & 53.2 \\
    \midrule
	SimCLR~\cite{chen2020simple} & R50-MLP & 200 & 61.9 & – & – \\
	MoCo v2~\cite{he2020momentum} & R50-MLP & 200 & 67.5 & ~84.0 & ~50.1\\
	PCL v2~\cite{li2020prototypical} & R50-MLP & 200  & 67.6 & 85.4 & 50.3\\
	\ProjectName~(Ours) & R50-MLP & 200  & \textbf{68.0}& \textbf{91.8} & \textbf{52.0} \\
\bottomrule
\end{tabular}
\end{center}
\end{table}

\subsection{Transfer to Object Detection}

We fine-tune the whole network following the experiment settings in detectron2, which are consistent with the other methods~\cite{he2020momentum,li2020prototypical}. Table~\ref{table: detection_coco} and Table~\ref{table: detection_voc} show that~\ProjectName~is overall better than MoCo v2 on COCO and VOC datasets.

\setlength{\tabcolsep}{4pt}
\begin{table}[t]
\begin{center}
\caption{Transfer learning results to object detection tasks on COCO dataset. The detection model is fine-tuned on COCO train2017 dataset and evaluated on COCO val2017 dataset.  All the baseline results are from~\cite{li2020prototypical}.}
\label{table: detection_coco}
\begin{tabular}{@{}l l l c c c c c c}
\toprule
    \multirow{2}{*}{Method}	 & \multirow{2}{*}{Arch} &pretrain  &\multicolumn{3}{c}{bbox} & \multicolumn{3}{c}{segm} \\
	& &epochs& AP & AP$_{50}$ & AP$_{75}$ & AP & AP$_{50}$ & AP$_{75}$ \\
\midrule
\noalign{\smallskip}
    Supervised & R50 & - & 40.0 & 59.9 & 43.1 & 34.7 & 56.5 & 36.9 \\
    \midrule
	MoCo v2~\cite{he2020momentum} & R50 & 200 & 40.7 & 60.5 & 44.1 & 35.4 & 57.3 & 37.6 \\
	\ProjectName~(Ours) & R50 & 200  & \textbf{40.8} & \textbf{60.6} & \textbf{44.3} & \textbf{35.5} & 57.3 & \textbf{38.0} \\
\bottomrule
\end{tabular}
\end{center}
\end{table}

\setlength{\tabcolsep}{4pt}
\begin{table}[t]
\begin{center}
\caption{Transfer learning results to object detection tasks on VOC dataset. The detection model is fine-tuned on VOC07+12 trainval dataset and evaluated on VOC07 test dataset. The baseline results are from~\cite{he2020momentum}.}
\label{table: detection_voc}
\begin{tabular}{@{}l l l c c c }
\toprule
    \multirow{2}{*}{Method}	 & \multirow{2}{*}{Architecture} &\#pretrain  &\multicolumn{3}{c}{VOC}\\
	& &epochs& AP$_{50}$ & AP & AP$_{75}$ \\
\midrule
\noalign{\smallskip}
    Supervised & R50 & - & 81.3 & 53.5 & 58.8 \\
    \midrule
	MoCo v2~\cite{he2020momentum} & R50 & 200 & 82.4 & \textbf{57.0} & 63.6 \\
	\ProjectName~(Ours) & R50 & 200  & \textbf{82.6}& 56.8 & \textbf{63.7} \\
\bottomrule
\end{tabular}
\end{center}
\end{table}

\section{Analysis}

The quantitative evaluation in Section~\ref{section:experiments} demonstrates that \ProjectName~not only outperforms other clustering methods on multiple benchmarks, but also improves the representation quality. Here, we further analyze the proposed objective to understand how it improves learning semantic and instance-wise information.

\subsection{Ablation studies}

$\mathbf{z}^c$ plays a crucial role in our method. We perform ablation studies on $\mathbf{z}^c$ to understand the importance of each technique. The evaluation results are reported in Table~\ref{table:ablations}, where the training is performed from scratch for 1200 epochs. First, we find that the lack of the equipartition constraint leads to a degenerate solution for cluster assignment, but has almost no effect on the training of representation learning. Second, we avoid the use of normalization on $\mathbf{z}^c$ and consider it as a regular logit like in classification. Our experiment shows that the loss becomes nan and the training fails, so the normalized $\mathbf{z}^c$ enables the corresponding dot product is bounded. Finally, we set the temperature $t$ to 1.0 (without $t$) and find that its value has an important impact on the clustering performance, which is analyzed in detail in Section~\ref{subsec:temperature}. We also train the model only using the equipartition constraint, and find that the model can not achieve optimal clustering results without the weighted InfoNCE loss. It also indicates that the weighted InfoNCE loss adjusts different weights of negatives to encourage $\mathbf{z}^c$ to capture more categorical information.

\setlength{\tabcolsep}{4pt}
\begin{table}
\begin{center}
\caption{Ablation studies of $\mathbf{z}^c$ on CIFAR10 including without the equipartition constraint, the normalization, temperature $t$ (default 1.0) and the InfoNCE loss. We compare the clustering results and linear evaluation on the pretrained backbone separately.}
\label{table:ablations}
\scriptsize
\begin{tabular}{@{}l cc@{}}
\toprule
Settings & Clustering (\%) & Linear evaluation (\%) \\
\midrule
Without constraint & 27.3 & 86.0 \\
Without norm & - & - \\
Without $t$ & 71.1 & 88.0 \\
Without InfoNCE & 67.0 & 86.0 \\
Full setup & \textbf{83.0} & \textbf{88.8} \\
\bottomrule
\end{tabular}
\end{center}
\end{table}
\setlength{\tabcolsep}{1.4pt}

\subsection{Temperature analysis}
\label{subsec:temperature}

Our method involves two temperatures: $\tau$ and $t$. Previous work~\cite{wang2021understanding} shows that temperature $\tau$ plays an important role in controlling the penalty strength of negative samples. Here, we consider $\tau$ as a constant and focus on the effect of $t$ on the clustering results, as shown in Figure~\ref{fig:temperature_clustering}. Since $t$ plays a role in the scaling of logits in the calculation of cross-entropy loss, a small $t$ reduces the difficulty of matching pseudo-assignment, and achieves better clustering results faster. In contrast, a larger $t$ value leads to the training of $\mathbf{z}^c$ becoming difficult. It is the presence of $t$ that balances the degree of difficulty between the instance-level discrimination and cluster assignment tasks. $t$ also allows $\mathbf{z}^c$ to be converted into proper softmax probabilities, as verified in the self-labeling experiments in Section~\ref{subsec:smaller_dataset}. The results demonstrate that similar clustering results can be achieved for a range of $t$, such as 0.1 or 0.2.

\setlength{\tabcolsep}{4pt}
\begin{table}
\begin{center}
\caption{The mean and standard deviation of similarity scores for $\mathbf{z}^c$ and $\mathbf{z}^n$ / $\mathbf{z}$ from a category perspective. Augmented: samples from the same instance, Same: samples from the same category, Different: samples from different categories. }
\label{table:z_statistics}
\begin{tabular}{lcccc}
\toprule
\multirow{2}{*}{Category}	 & MoCo v2 & &\multicolumn{2}{c}{\ProjectName}\\
\cmidrule{2-2} \cmidrule{4-5}
& $\mathbf{z}$ & & $\mathbf{z}^c$ & $\mathbf{z}^n$ \\
\midrule

Augmented  & 0.781 $\pm$ 0.188 & & 0.872 $\pm$ 0.154 & 0.720 $\pm$ 0.199 \\
Same  & 0.062 $\pm$ 0.175 & & 0.504 $\pm$ 0.237 & 0.004 $\pm$ 0.183 \\
Different & -0.006 $\pm$ 0.115 & & -0.033 $\pm$ 0.323 & -0.001 $\pm$ 0.167 \\
\bottomrule
\end{tabular}
\end{center}
\end{table}
\setlength{\tabcolsep}{1.4pt}

\begin{figure*}[!t]
\vskip 0.2in
  \centering
  \includegraphics[width=0.8\textwidth]{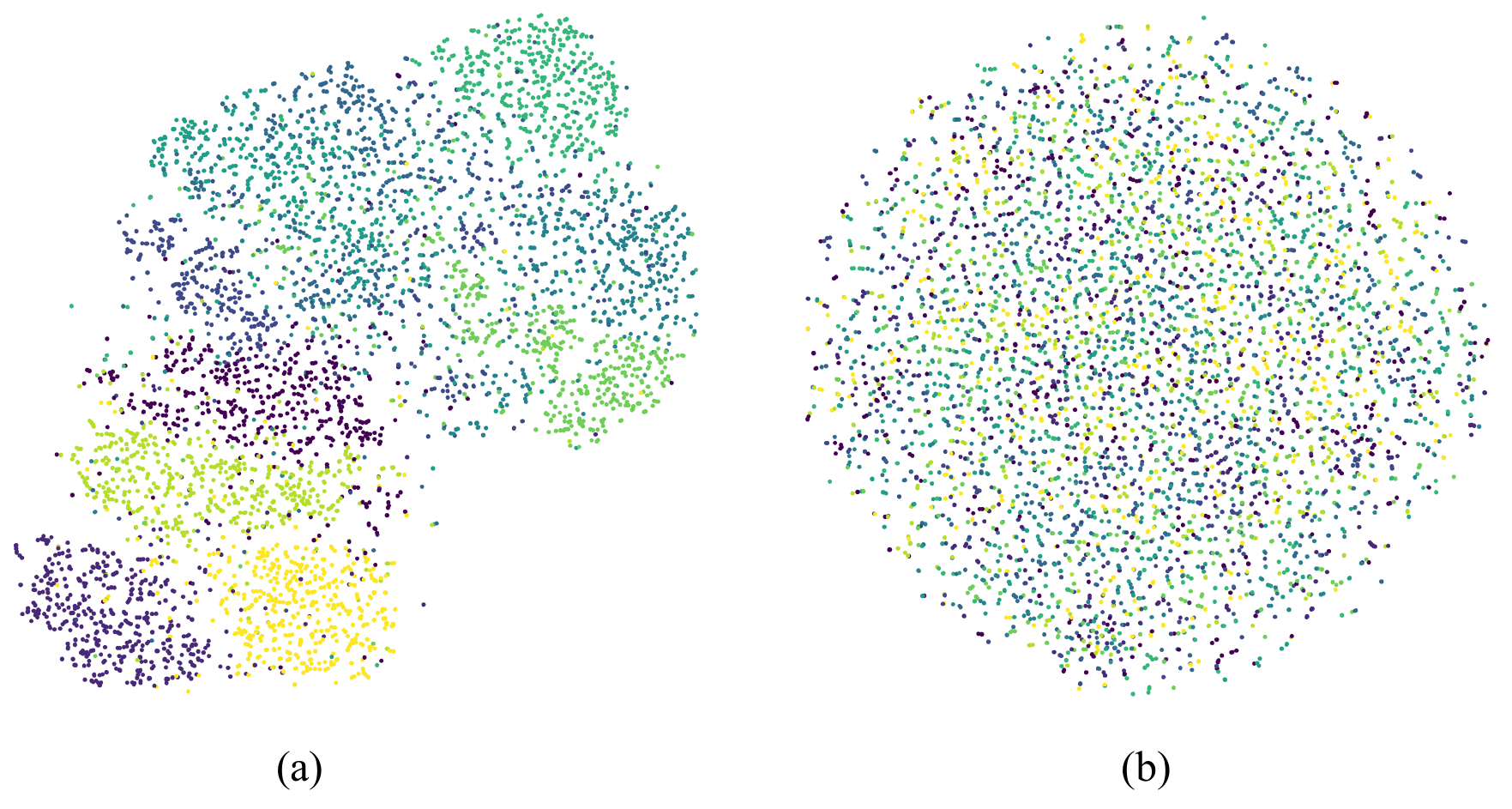}
  \caption{The t-SNE visualization of $\mathbf{z}$ / $\mathbf{z}^n$ on CIFAR10 test dataset. (a): MoCo v2, (b): \ProjectName(ours). Colors indicate different categories.}
  \label{fig:zn_visualization}
\vskip -0.2in
\end{figure*}

\subsection{Latent space analysis}
\label{subsec:latent}

Table~\ref{table:z_statistics} shows the statistics of similarity scores (dot product), where MoCo v2 has only $\mathbf{z}$ to compute the contrastive loss, and our method has $\mathbf{z}^c$ and $\mathbf{z}^n$. Both methods satisfy the requirements of instance-wise contrastive learning well, where positive samples (Augmented) have high similarity scores of $\mathbf{z}$ / $\mathbf{z}^n$ and negative samples (Same or Different) have low similarity scores of $\mathbf{z}$ / $\mathbf{z}^n$. In~\ProjectName, since $\mathbf{z}^c$ encodes the categorical information, the similarity score of $\mathbf{z}^c$ is also able to distinguish well between samples from different categories. As analyzed in Section~\ref{subsec:ICL}, it provides a weight adjustment mechanism for different negative samples so that it can handle negative samples of different hardness well. In contrast, MoCo sets the weight of all negative samples to 1, which tends to learn larger dot products for positive samples. The previous study~\cite{wang2020understanding} summarizes two key properties of contrastive loss: alignment and uniformity. In other words, MoCo is more concerned with alignment for the optimization purpose.

We further analyze the uniformity properties of $\mathbf{z}$ / $\mathbf{z}^n$ of MoCo and~\ProjectName~using t-SNE~\cite{van2008visualizing} and the results are shown in Figure~\ref{fig:zn_visualization}. Compared to $\mathbf{z}$ learned by MoCo, our method tends to uniformly distribute points over the latent space without preserving any category-related information. Although MoCo achieves instance-level differentiation, we can still observe that points of the same category are clustered together. The main reason is that the typical contrastive loss cannot deal with the hard negative problem well and samples of the same category aren't distributed evenly. And the proposed method enables us to learn a uniformly distributed space due to the mechanism of self-adjusting negative weights. This also shows that the MLP projection head in our method works well as the role of transforming in
two representation spaces. Therefore, we decompose the original $\mathbf{z}$ into two separate parts: $\mathbf{z}^c$ related to the categorical information, thus focusing on clustering, and $\mathbf{z}^n$ related to instance-wise information, thus focusing on alignment and uniformity. Our method also demonstrates that the MLP projection head plays a role in the transformation from the linearly separable feature space to the instance-wise representation space.

\section{Related work}


\textbf{Disentanglement.} Learning disentangled representations can be used to uncover the underlying variation factors within the data~\cite{bengio2013representation}. Generally, existing disentangling methods can be classified into two distinct categories: The first category involves separating the latent representations into two~\cite{mathieu2016disentangling,zheng2019disentangling,patacchiola2020autoencoders} or three~\cite{gonzalez2018image} parts. For instance, Mathieu \etal~\cite{mathieu2016disentangling} introduced a conditional variational autoencoder (VAE) combined with adversarial training to disentangle the latent representations into factors relevant to labels and unspecified factors. In these two-step disentanglement methods~\cite{zheng2019disentangling}, the initial step involves extracting label-relevant representations by training a classifier, followed by obtaining label-irrelevant representations primarily through reconstruction loss. These approaches aim to enhance disentanglement performance by leveraging (partial) label information and minimizing cross-entropy loss. The second disentanglement category, exemplified by methods like $\beta$-VAE~\cite{higgins2016beta}, FactorVAE~\cite{kim2018disentangling}, and $\beta$-TCVAE~\cite{chen2018isolating}, focuses on learning to disentangle each dimension in the latent space without manual annotations. Unlike many disentanglement learning methods (such as those based on generative models like VAEs and GANs), which have been proposed in previous studies~\cite{patacchiola2020autoencoders,dupont2018learning,ding2022clustering}, this paper introduces a novel approach. The proposed method utilizes contrastive learning to train the encoder and achieves the disentanglement of latent variables.

\textbf{Clustering.} Recent research has focused on the joint learning of feature representation and clustering, either in an alternating or simultaneous manner. Earlier studies (\eg~IMSAT~\cite{hu2017learning}, IIC~\cite{ji2019invariant}) focus on learning a clustering method from a mutual information maximization perspective. Since these methods may only focus on low-level features, such as color and texture, they don't achieve excellent clustering results. DeepCluster~\cite{caron2018deep} performs clustering and representation learning alternatingly, which is further improved with online clustering~\cite{zhan2020online}. Self-labelling~\cite{asano2019self} is another simultaneous learning method by maximizing the information between labels and data indices. ContrastiveClustering~\cite{li2021contrastive} proposes two MLP heads and handles instance- and cluster-level contrastive learning separately, which results in the use of two distinct contrastive losses. In this work, we propose a self-adjusting contrastive loss to directly disentangle the feature representation for clustering.

\textbf{Contrastive learning.} The instance-wise contrastive learning considers each sample as its own class. Wu \etal~\cite{wu2018unsupervised} first propose to utilize a memory bank and the noise contrastive estimation for learning. He \etal~\cite{he2020momentum} continue to improve the training strategy by introducing momentum updates. SimCLR~\cite{chen2020simple} is another representative work that applies a large batch instead of the memory bank. SwAV~\cite{caron2020unsupervised} and PCL~\cite{li2020prototypical} bridges contrastive learning with clustering to improve representation learning. There are also some recent works~\cite{grill2020bootstrap,chen2021exploring,caron2021emerging} that consider only the similarity between positive samples. Although the typical contrastive loss enables learning a latent space where each instance is well-differentiated, it cannot deal well with the uniformity of the hard negative samples. In our work, the proposed contrastive loss with self-adjusting negative weights solves this problem well. Once the feature representation is obtained, cluster assignment is either obtained by K-means clustering~\cite{xie2016unsupervised,caron2018deep,li2020prototypical} or training an additional component~\cite{ghasedi2017deep,van2020scan}. However, they still cannot achieve promising clustering results on a large-scale dataset, \eg, ImageNet, which requires to develop an efficient objective function that jointly learns cluster assignment and representation learning.

\section{Conclusions}

This work introduces a novel approach to learning disentangled representations for clustering using contrastive learning. The proposed method decomposes the representation into two distinct parts: one dedicated to clustering and the other focused on instance-wise learning. This enables end-to-end training, distinguishing it from existing state-of-the-art clustering methods that typically involve two separate steps, where representation learning and clustering are decoupled. By avoiding this decoupling, our method has the potential to achieve superior results, especially on large-scale datasets. Experiments on multiple benchmarks demonstrate that our disentanglement method not only achieves excellent clustering performance, but also improves contrastive learning. Note that \ProjectName~can be combined with over-clustering, vision transformers, advanced augmentation and training strategies. Due to the limitations of our computational resources, we will explore these techniques in future work.

\section{Acknowledgments}

The work of Feng Luo was supported in part by the U.S. National Science Foundation (NSF) under Grant ABI-1759856, Grant MRI-2018069, and Grant MTM2-2025541, the U.S. National Institute of Food and Agriculture (NIFA) under Grant 2017-70016-26051.


%






\ifCLASSOPTIONcaptionsoff
  \newpage
\fi



\bibliographystyle{IEEEtran}
\bibliography{references}
%

%

%


\newpage

\begin{IEEEbiography}
[{\includegraphics[width=1in,height=1.25in,clip,keepaspectratio]{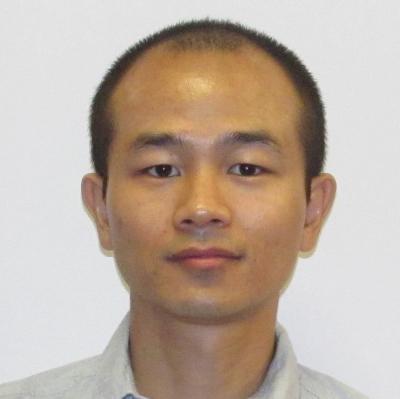}}]{Fei Ding}
is currently pursuing the Ph.D. degree with the School of Computing, Clemson University. His research interests include deep neural networks, unsupervised learning and machine learning.
\end{IEEEbiography}

\begin{IEEEbiography}
[{\includegraphics[width=1in,height=1.25in,clip,keepaspectratio]{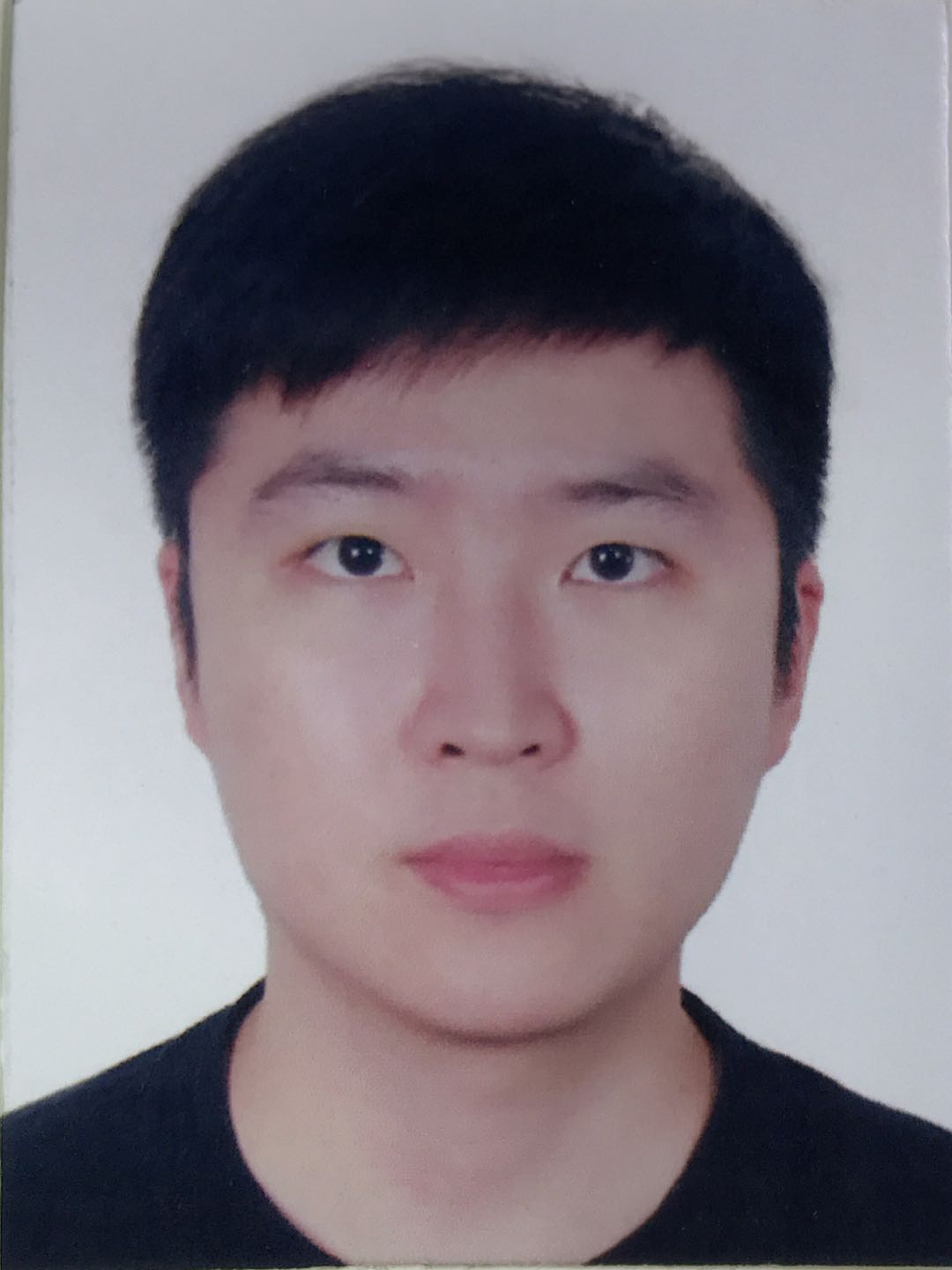}}]{Dan Zhang} received the B.E. degree from Sichuan University, China, in 2012 and the M.S. in computer science from New York University in 2015. His research interests include Deep Learning, Self-supervised Learning and generative model in computer vision.
\end{IEEEbiography}

\begin{IEEEbiography}
[{\includegraphics[width=1in,height=1.25in,clip,keepaspectratio]{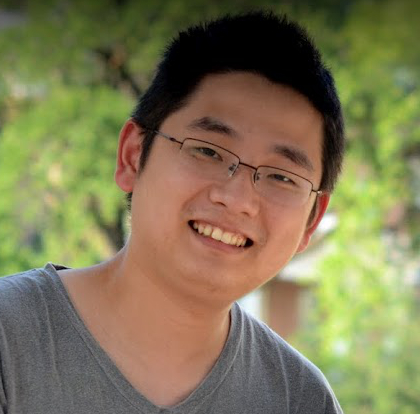}}]{Yin Yang}
(Member, IEEE) received his Ph.D. degree in computer science from the University of Texas at Dallas in 2013. He is currently an associate professor with School of Computing, Clemson University. He is a recipient of the NSF CRII award (2015) and CAREER award (2019). His research aims to develop efficient and customized computing methods for challenging problems in Graphics, Simulation, Machine Learning, Vision, Visualization, Robotics, Medicine, and many other applied areas.
\end{IEEEbiography}

\begin{IEEEbiography}
[{\includegraphics[width=1in,height=1.25in,clip,keepaspectratio]{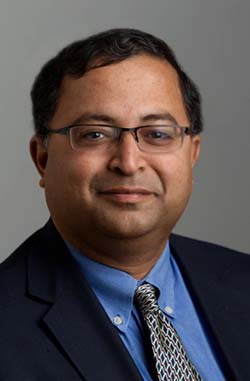}}]{Venkat Krovi}
received his Ph.D. degree in Mechanical Engineering and Applied Mechanics from the University of Pennsylvania in 1998. He is the Michelin Endowed Chair of Vehicle Automation in the Departments of Automotive Engineering and Mechanical Engineering at Clemson University. His research activities include plant-automation, consumer electronics, automobile, defense and healthcare/surgical simulation arenas. His work has been published in more than 200 journal/conference articles and book chapters, and patents.

\end{IEEEbiography}

\begin{IEEEbiography}
[{\includegraphics[width=1in,height=1.25in,clip,keepaspectratio]{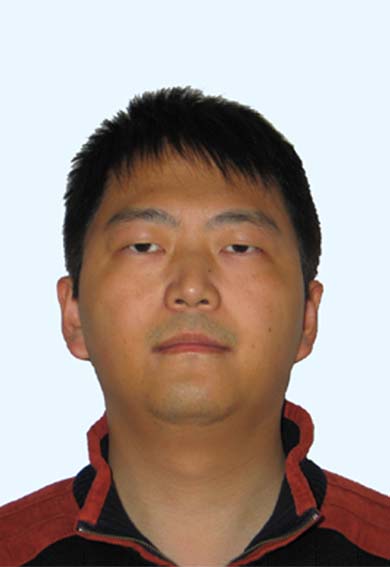}}]{Feng Luo}
(Senior Member, IEEE) received the Ph.D. degree in computer science from The University of Texas at Dallas, in 2004. He is the Marvin J. Pinson, Jr. ’46 Distinguished professor of School of Computing, Clemson University and the founding director of Clemson AI Research Institute for Science and Engineering (AIRISE). His research interests include deep learning and application, high throughput biological data analysis, data-intensive bioinformatics, network biology, and computational genomics and genetics.
\end{IEEEbiography}

\end{document}